# Measuring Similarity: Computationally Reproducing the Scholar's Interests

Ashley Lee[1], Jo Guldi[2], Andras Zsom[1]

Computerized document classification already orders the news articles that Apple's "News" app or Google's "personalized search" feature groups together to match a reader's interests. The invisible and therefore illegible decisions that go into these tailored searches have been the subject of a critique by scholars who emphasize that our intelligence about documents is only as good as our ability to understand the criteria of search (Hitchcock, 2013). This article will attempt to unpack the procedures used in computational classification of texts, translating them into term legible to humanists, and examining opportunities to render the computational text classification process subject to expert critique and improvement.

This article describes a process for measuring similarity between two sets of documents – one a large-scale corpus, the other a set of known secondary sources – using a combination of divergence measurements, expert reading, and sampling (see Figure 1).

[1] Brown University, CIS, Data Science Practice
[2] Southern Methodist University, History Department

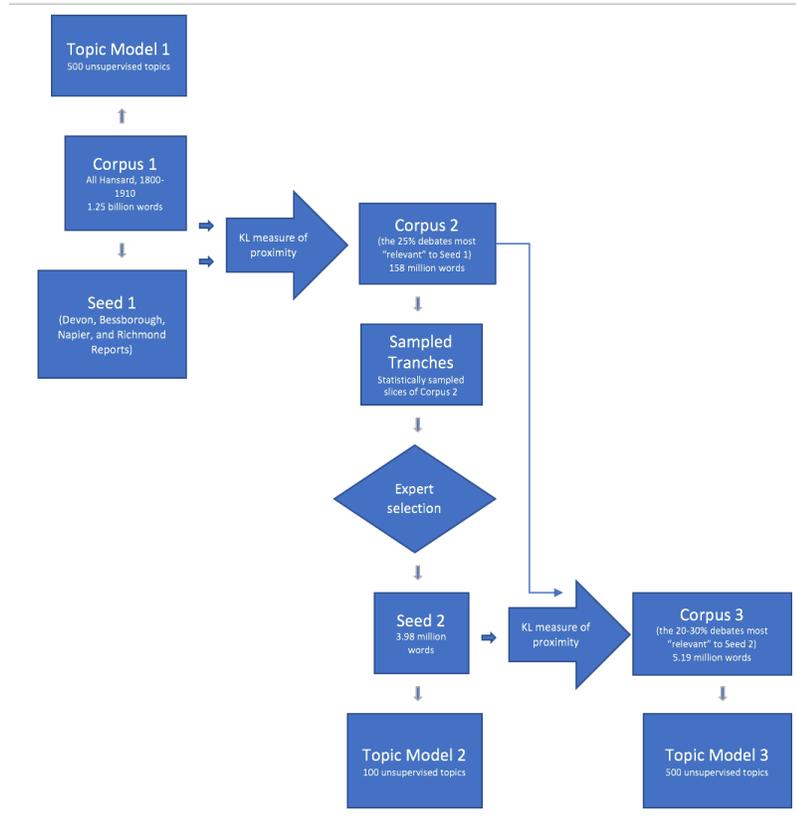

Figure 1: The Proposed Process

A first layer of analysis consisted in a computer-aided search based on the match between Hansard debates and our "seed texts," producing Corpus 2, a subset of Hansard texts more likely to concern the subject of eviction and agrarian crisis.

The four texts of Seed 1 are notable as parliamentary reports that concern eviction and agrarian crisis: the Devon Report that followed the Irish famine in 1848; the Bessborough Report of 1881, issued in the midst of the Irish Land War; the 1884 Napier Report generated by a commission that interviewed hundreds of Scottish Crofters; and the Richmond Report of 1881, which examined English agrarian uprisings against an international study in how governments could support agricultural prices. As parliamentary reports drafted by members of parliament in committee, they share with parliamentary debates the language of the period and the idiom of reform, and therefore seemed to offer a suitable baseline for matching other discussions of eviction and agricultural tenancy. The reports cluster around 1881 with the exception of the Devon Report (1845); and this clustering in time represents one weakness of a baseline that should be designed to match discussions of property across the century. However, it was felt that the temporal weakness was compensated by the length of the Devon Report (at nearly 1000 pages) against the much shorter Bessborough Report (11 pages plus evidence) and the opposite political positions taken in the pro-landlord Devon Report and the land-reform Bessborough and Napier reports. These texts are designed to provide a broad overview of debates about rent, leasing, titling, property registration, landlord profitability, eviction, tenant rights, commonage rights, trespass, infrastructure, improvement, and other subjects that would be closely debated in any parliamentary inquiry into property.

In the winnowing process that followed, the entire Hansard Corpus was cut down to a quarter of its size. The similarity algorithm was used to match the 25% of the Hansard corpus most similar to the seed texts about property. Corpus 1 (Hansard as a whole from 1800-1910) has 111,685 speeches and a quarter billion words. Corpus 2 has 27,921 debates and 158 million words. Topic Model 2 summarizes the contents of this selection in 100 topics.

Corpus 2 was then subjected to sampling and human analysis in a second round of analysis. In the machine learning literature, the results of scholarly inspection are considered a "validation sample," which can be used to tune the parameters of an algorithm (Mohri et al). This process determined that roughly 20% of Corpus 2 was most relevant to research questions about property. Another round of winnowing was applied. A similarity algorithm was used to match the 25% of Corpus 2 that was most similar to the results of the validation sample (Seed 2). The result was a final corpus of 1,396 debates or 5.19 million words. Topic Model 3 summarizes the contents of this selection.

At each stage in the winnowing process, a topic model was used to check kinds of documents comprise the results. In keeping with scholarly work in the digital humanities, the topics were assigned names based on their most statistically significant words (Blevins, 2011). Finally, we winnowed Corpus 2 through a further unsupervised matching process, using Seed 2 as the basis for a similarity search, the result of which is labeled Corpus 3. Corpus 3 was then the basis for a further round of analysis that produced Topic Model 3. These topics were also assigned names. It is to the interpretation of those resulting topics to which we now turn.

*The Mathematical Basis for the Process: Similarity*

Measures of approximative similarity between different probability distributions is where computation and math excel. After we have broken up our texts into words, the next step is to compare the seed texts and Hansard so as to find the debates most similar to the seed. In order to execute the comparison in such a way as to preserve the co-occurrences of words within individual speeches, we must choose *a distance metric*, which assigns a measure to every debate in Hansard of how "similar" that debate is to the seed topics. Measuring distance will allow us to begin with the seed reports in which parliament talks about eviction, and then to derive a subcorpus of speeches that meet a mathematical threshold of similarity to the seed corpus.

Our chosen method of pulling out a derived corpus is based on *divergence*, a subset of distance measures which concentrates on measuring the similarity between two strings of information. Divergence makes sense, in part, because we are measuring strings of text rather than individual numbers that can be easily subtracted from each other. Divergence measures the similarity between two strings or distributions, producing another distribution that is interpreted as the distance between the two.

Divergence between the seed texts and Hansard as a whole can be represented as a histogram of speeches each of which has a different distance from the seed text. The resulting distribution groups the most similar speeches to the left of the x axis, and the least similar to the right of the x axis. The y axis tells us how many speeches are in each category: a few are very similar to the seed texts, many are somewhere in between, and a few are very far away.

We are in good company in our choice. KLD is routinely used in digital humanities as a means to classify texts, including to rank documents' *proximity to topics* discovered in topic modeling, or to rank nouns and verbs that regularly occur together (Bigi, 2003).

Understanding the difference between possible measures that can be used to create the divergence in question is crucial. While mathematically similar, our preliminary results suggest that different measures of divergence will produce an utterly different sample of historical texts. For instance, compare the top 1% of similar documents over time returned by three different divergent measures in the figure below.

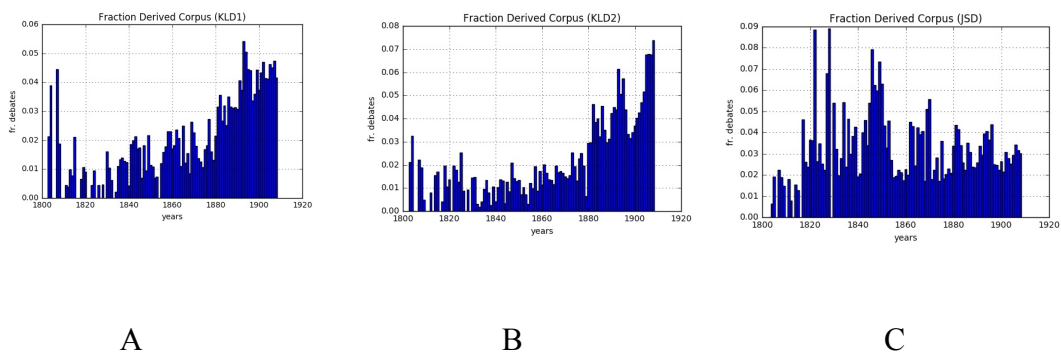

|  A  |  B  |  C  |

*Figure: Documents Most Similar to the Seed Documents (the top 1%) Over Time, according to (a) Kullback-Leibler Divergence, (b) Symmetrical Kullback-Leibler Divergence, and (c) Jenson-Shannon Divergence.*

Were a scholar relying purely on a measure of divergence for an account of how much property was debated in Parliament, she would come away with a very different account depending on the measure she used: with the first two measures, she would imagine that property was debated increasingly over the century; where with the third measure, she would imagine that debates about property peaked in the 1840s and 50s and subsequently declined. If the outcome between different mathematical metrics of distance can offer such different historical conclusions, we should be very careful indeed in our choice of measures.

An intelligent choice should reflect an understanding of how, in each case, information and noise are related to each other. The principle differences between the measures are *symmetricality*, which has to do with the relationship between the strings that one is comparing. These strings may be conceived of as equally valid patterns that have some concrete relationship with each other (symmetrical), or the strings may be one pattern of noise and another pattern of truth (asymmetrical).

For our work, the relationship between the two strings under comparison can be understood as that between a *seed* -- the parliamentary reports previously cited in the scholarly literature -- and a *full corpus*, the full text of the Hansard debates. Both are equally valid corpora for study in terms of scholarly interest in general, but if we are looking for questions relating to property, only the *seed* text constitutes an object of interest; for a historian of property, the full text of Hansard is comprised of noise that swirls into different shapes -- here about abolition, there about the coronation -- none of them having any relationship to the shape of the property corpus. Mathematically speaking, the seed reports are our "true distribution" - they contain language about land and property that we know we are interested in. We ultimately opted for a non-symmetrical KLD.

## *Choice of Object: Distance Between Bags of Words*

There are different procedures for mathematically measuring similarity in a textual corpus. Some scholars classify similar texts together first using vectors; others use matrices, and still others use probability. All three mathematical models deal in cutting up texts into words, the so-called "bag-of-words" model that eliminates grammar; prominent words in the seed texts include *tenant, landlord, land*, and *rent*. In Hansard overall, prominent keywords include *reform, honor, friend, speech,* and *bill*. We will create a distribution for the words in the seed texts, and another distribution for the words in each individual speech, whether it concerns the abolition of slavery or the coronation of Victoria.

| Rank | Word | Number of Occurrences |
|---|---|---|
| 1 | rent | 6686 |
| 2 | land | 6103 |
| 3 | tenants | 5748 |
| 4 | tenant | 5155 |
| 5 | landlord | 4045 |
| 6 | case | 3383 |
| 7 | estate | 3094 |
| 8 | farm | 2843 |
| 9 | made | 2505 |
| 10 | time | 2464 |
| 11 | mr | 2292 |
| 12 | pay | 2275 |
| 13 | property | 2116 |
| 14 | cases | 2085 |

| 15 | year | 2074 |
| 16 | lease | 1943 |
| 17 | rents | 1843 |
| 18 | country | 1820 |
| 19 | valuation | 1798 |
| 20 | man | 1785 |

*Table: Top N-Grams, Seed Corpus*

What does a probability model based on the "bag of words" model look like? We can experiment by breaking up our seed texts into individual words, organized by the probability of their co-occurrence into categories known as "topics." For our seed corpus, an initial review suggests that topic modeling has correctly identified some of the themes of interest to property scholars. For example, the most prevalent topic overall, which we might name as a "Property Negotiation," contains words relevant to the negotiation of land and rent prices, including the terms "tenant," "landlord," "give," "pay," "case," "agent," "sell," "refus*" (the * here is a wildcard, lumping together "refusal" with "refusing," etc.), "farm," "reason," "object," and agre*." Other top topics similarly evidence that abstraction into topics reduplicates our scholarly concerns:

| Scholar-assigned name of the topic | Relative prevalence of the topic amongst the seed texts | Top words in the topic (as grouped by computer probability measure) |
|---|---|---|
| "Property negotiation" | 0.04249 | "tenant","hold","farm","small","part","land","properti","tenants","purchas","sold","estat","chang","occupi","holdings","case","anoth","portion","ireland","owner","bought","" |
| "Property exchange" | 0.03585, | "tenant","hold","farm","small","part","land","properti","tenants","purchas","sold","estat","chang","occupi","holdings","case","anoth","portion","ireland","owner","bought","" |
| "Rental history" | 0.03566 | "rent","pay","year","rais","paid","increas","due","arrear","rise","half","cent","sinc","put","raised","high","payment","chang","charg","case","made","" |
| "Tenant right" | 0.03378 | "tenant","landlord","interest","sell","pay","purchas","power","hold","properti","tenantright","contract","sale","put","tenants","becom","paid","owner","make","subject", |

|  |  | "charg", |
|  |  |  |
| "Improvement" | 0.03192 | 0.03235,"countri","present","state","condit","part","peopl","country","general","system","district","improv","live","natur","labour","ireland","found","posit","people","circumst","practic","" |

*Table: Top topics in the Seed Texts (as measured by Latent Dirichlet Analysis)*

The groupings, we must insist, reflect frequent combinations found in the text themselves. They are not scholarly-assigned; they are not, for example, the result of my matching words similar to "tenant" using a thesaurus. Note that "tenant" appears in three of the five top topics from the seed texts. The topics are not exclusive groupings; they are probablistic guesses at how likely certain words are to co-occur with each other.

## *Results: Different Similarities*

To explore the results of our measurements, we displayed the measure of difference between each seed text and the corpus as a whole as bell-curve distributions of divergence between the Hansard debates and seed texts, where the x axis is a scale of similarity from least distant to most distant, and the y axis is the number of debates at each level of similarity (see Figure).

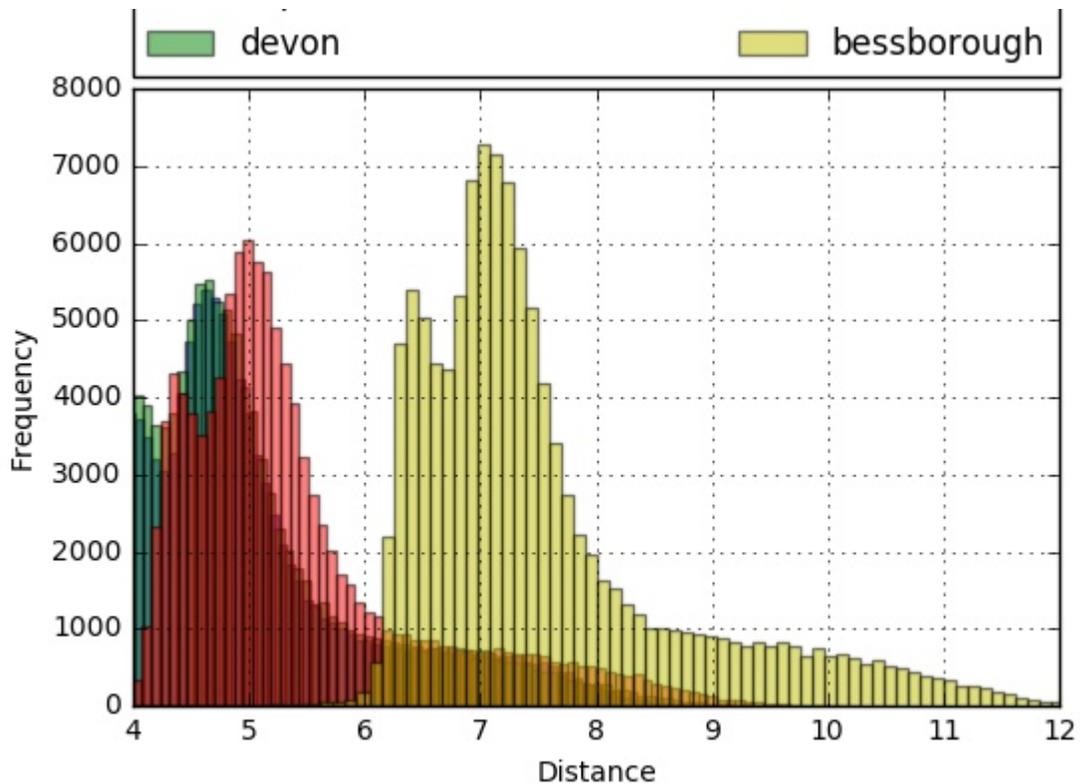

*Figure: Similarity between each seed report (by color) and the Hansard debates*

The bell-curve-like shape results suggesting that some debates (the short tail towards the left of the chart) are quite similar to each seed text, while a vast body are loosely related to themes like "improvement," and a tiny number are barely related to this language at all. The bell curve, if we trust it at all, gives us an indication of where we might start reading -- at the leftmost part of the graph, where a few key debates have been measured to most closely match our seed topics.

Having defined similarity and asked the computer to extract the texts most similar to our seed corpus, we will next use the debates at the left end of the bell curve to check what's in the derived corpus. After all, a scholar must be sure that she has the correct subset of Hansard debates for her work in order that analysis of the subcorpus should be meaningful.

## *Expert Verification*

As part of refining the process of research, we asked an expert to guide by reviewing the individual texts labeled "similar" by the computer. Our procedure therefore calls for a round of interpretation. For our immediate purposes, we want to read several tranches of similar documents so that we can choose a new set of "relevant" debates that we will use to. From these tranches, we will iteratively sample the most similar texts to the seeds so that we can "train" our process. We may engage in two or three rounds of iterative sampling. The iterative process of sampling, expert classifying, and training the algorithm constitutes our method of refining the computer's similarity ranking using new information. The sorting of the computational category of "similar" from the scholar's category of "relevant" can only be done by reading, although we are encouraging a model of reading "samples" rather than reading the whole corpus in question.

Here, our method takes a detour from automatic, mathematical detection into critical thinking about what happens when the computer's match does not mirror what the scholar would have done. Ideally, using our parliamentary reports about eviction, the computer will match Hansard debates where eviction, tenants, and landlords come up regularly. You can read the titles of the debates that the computer matched with our seed corpus here (there are three different texts corresponding to three different ways of measuring what's "similar" -- we will define those methods in our terms, below). When you do, you will probably be appalled at the entire process we've put you through thus far.

The initial results of the derived corpus process were far from an obvious match. The top 5% matched documents are an unwieldy tour of Hansard. Some of them match our concerns with eviction and tenants -- a promising 1835 debate on the "Law of Landlord and Tenant (Ireland)", an 1845 debate on the Thames Embankment, but many more of them match our research interest very little: A "Mutiny Bill" on flogging in the army in 1827; "Duties on Wool" in 1830; an 1842 debate on the "Printing Committee" that circulated parliamentary agendas and reports. This selection clearly won't do as a process of selecting documents. A sample and review concludes that the "hit rate" similar texts that were relevant to scholarly interest was quite low; only 22% of the texts at the 5% cutoff were indeed classifiable as "about property."

In a traditional research process, the scholar already expects to refine her questions and to limit her reading material after first consulting a catalog. Just so, we recommend first using a KL similarity measure to guide wide reading, and using the results of that wide reading to narrow the search progressively.

## *Second Result: Seed 2*

These result of this process – Seed 2 – have a different chronology than the first set of texts, with a majority located after 1881.

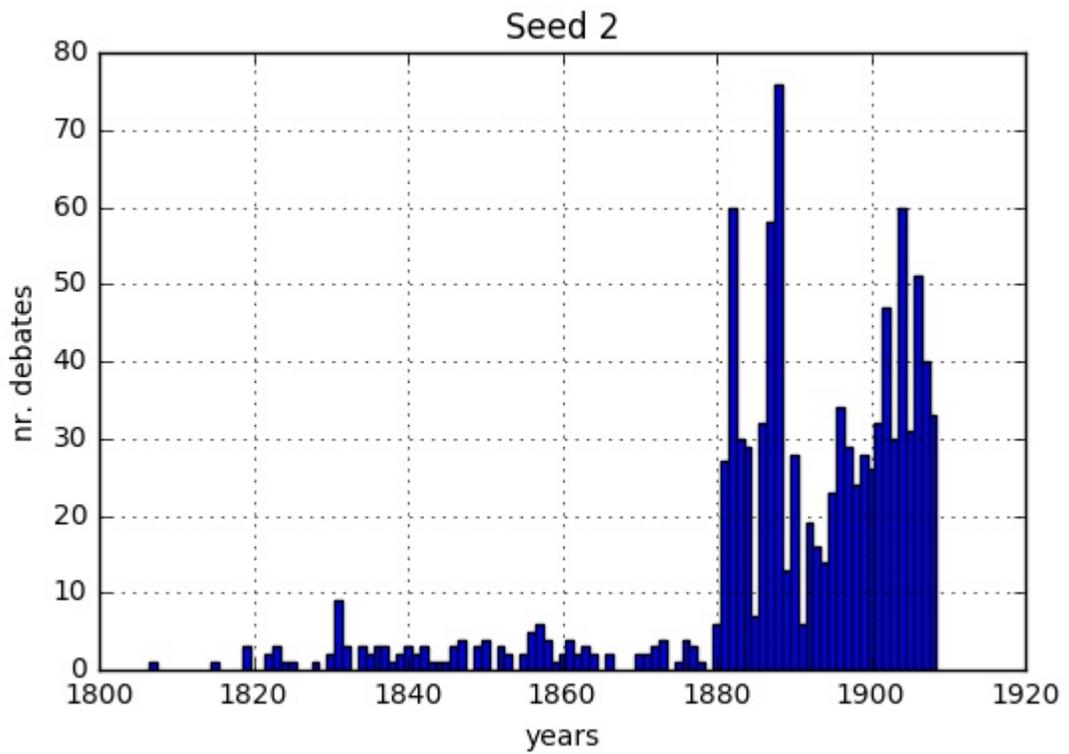

*Figure: Histogram of the Second-Round Hansard Debates Chosen by the Expert for their Relevance, Showing Frequency by Year*

The resulting texts of Seed 2 are also evidence more specificity in the range of topics the cover, when the texts are modeled by a topic-modeling algorithm (see table).

| Scholar-assigned name of the topic | Relative prevalence of the topic amongst the seed texts | Top words in the topic (as grouped by computer probability measure) |
|---|---|---|
| Interrogation of the evidence in liberal reform | 0.29357 | question opinion refer larg purpos difficulti obtain express put brought circumst practic desir proper mani agre district condit subject accord |
| State methods of gathering evidence | 0.26255 | receiv question matter state made inform report inquiri inquir make respect recent held refer order term repli result caus |
| Colonial government in Ireland | 0.25249 | lord ireland secretari chief lieuten attent counti fact polic call propos step govern direct o'connor place hous sir trevelyan michael |
| Legislative governance of Ireland by parliament | 0.28143 | act land section made ireland tenant commiss provis claus applic work secretari provid treasuri appli rule power law hold view |
| Parliamentary examination of colonial rule in Ireland | 0.22295 | land awar beg delay made step ireland arrang longford part state reason counti inspect caus prevent sinc enter repres date |

| Information about land made available for political purposes | 0.21586 | return land inform report refer give total made obtain laid lay number estim tabl grant form statement annual amount show |
| --- | --- | --- |
| Oversight of the land commission | 0.20721 | secretari ireland chief lord lieuten beg commission balfour cork awar land paragraph irish gerald commiss manchester brought morley commissioners assist |
| Tenant purchase of land under the Land Commission | 0.16621 | purchas land sale advanc commission tenant interest hold money year agreement amount farm price sanction irish agre valu number requir |
| Grounds for government intervention include feelings, petition, and testimony | 0.10939 | made case hous general house long caus person hand alway give moment **feel** natur mind bring refus submit interfer noth |
| The expense of reform | 0.06689 | cost case rule awar consent year expens parti alter continu entitl increas born suggest advis adjust parties adopt run costs |
| Judicial intervention | 0.03299 | court judg order case colonel lloyd shaw copi appeal judgment petti justic ann turf brought decre cut poor curran conduct |
| Land disputes in Limerick, | 0.03239 | farm leitrim limerick agent effect sheriff patrick |
| Leitrim, and Roscommon | | roscommon car approv messrs kingston till o'kelli evid landlord owen **hewson** sell defend |
| Land disputes in Donegal | 0.02189 | offer sold estate lot forc donegal shirley receiv olphert tenantri necessari adjourn monaghan crown lifford repres chief falcarragh withhold telegram |
| The social contract of the rents will be mediated by the courts | 0.01875 | gentleman peopl govern learn statement brought civil execut government rents night cours applic legal vindic recov illeg offenc feel extract |
| Legislative intervention | 0.0116 | hous landlord govern amend amendment word crown deal condit perfect justifi island sympathi accept secretari forc militari action refus power |
| Victims of eviction | 0.01113 | land acr scotland **pollok** mountain improv depopul side class heath arabl inhabit capabl friend import acres entir thought product turn |
| Settler colonialism in Canada and Patagonia | 0.00906 | foreign canada welsh colonist affairs investig thoma chubut deleg slatteri free scott adjourn canadian patagonia tydvil merthyr david denni coloni |
| Land disputes in Skye | 0.00491 | skye sheriff expedit case scotland arrear rates prison militari offic justic person due polic tri |

|  |  | inquiri ivori inquiry island made |
| --- | --- | --- |
| Forest land disputes in Scotland | 0.00495 | deer forest scotland peopl nobl year enjoy sheep forests exclus highlands natur petit visitor sport extent air stranger game anim |
| Access to mountains in Scotland | 0.0049 | mountain tourist sport hill aberdeen harm legisl prevent braemar interdict propos access walk exclud neighbourhood danger amount lamb uncultiv highland |

*Table: Major topics and interesting topics from the expert-picked sample of debates[3]*

As a result of these findings, we propose that measuring divergence, with additional input by the author, offers an important tool for researchers who wish to use tools to find similar documents.

## *References*

Tim Hitchcock, "Confronting the Digital: Or How Academic History Writing Lost the Plot," Cultural and Social History 10, no. 1 (2013): 9–23.

Mohri, Rostamizadeh, and Talwalkar, Foundations of Machine Learning (Cambridge: The MIT Press, 2012).

---

[3] This particular selection of topics was run with stemming, rather than lemmatizing, which we would soon correct in our process. Stemming breaks words arbitrarily at a "stem" that the computer presumes may be the root of similar words, and results in the aggregation of "practical" and "practicing," which the computer treats as the same word despite the fact that they have different meanings. Lemmatizing uses grammatical information to class "buy" and "bought" as the same word, even though they do not share a common stem. In later versions of our topic work we corrected the model. However, with revised input, it was unlikely that the same proper nouns would come up as they are presented here, so we have opted to present the initial returns.